\documentclass[journal]{IEEEtran}
\ifCLASSINFOpdf
\else
\fi

\usepackage{cite}
\usepackage{url}
\usepackage{amsmath}
\usepackage{amssymb}
\usepackage{hyperref}
\usepackage{multirow}
\usepackage{graphicx}
\usepackage[dvipsnames]{xcolor}

\begin{document}
%
%
\title{Template-Free Try-on Image Synthesis via Semantic-guided Optimization}
%
%

\author{Chien-Lung Chou, Chieh-Yun Chen, Chia-Wei Hsieh, Hong-Han Shuai, Jiaying Liu, and Wen-Huang Cheng
\IEEEcompsocitemizethanks{
\IEEEcompsocthanksitem C.-L.~Chou, C.-W.~Hsieh, and H.-H.~Shuai are with the Department of Electrical and Computer Engineering, National Chiao Tung University. 
E-mail: \{chienlung.eed04,maggie1209.tem04,hhshuai\}@nctu.edu.tw.
\IEEEcompsocthanksitem C.-Y.~Chen and W.-H.~Cheng are with the Institute of Electronics, National Chiao Tung University. 
W.-H.~Cheng is also with the Artificial Intelligence and Data Science Program, National Chung Hsing University. 
E-mail: \{cychen.ee09g,whcheng\}@nctu.edu.tw.
\IEEEcompsocthanksitem Jiaying Liu is with the Wangxuan Institute of Computer Technology, Peking University. 
E-mail: liujiaying@pku.edu.cn.
}
}

\maketitle

\begin{abstract}
The virtual try-on task is so attractive that it has drawn considerable attention in the field of computer vision. However, presenting the three-dimensional (3D) physical characteristic (e.g., pleat and shadow) based on a 2D image is very challenging. Although there have been several previous studies on 2D-based virtual try-on work, most 1) required user-specified target poses that are not user-friendly and may not be the best for the target clothing, and 2) failed to address some problematic cases, including facial details, clothing wrinkles and body occlusions. To address these two challenges, in this paper, we propose an innovative template-free try-on image synthesis (TF-TIS) network. The TF-TIS first synthesizes the target pose according to the user-specified in-shop clothing. Afterward, given an in-shop clothing image, a user image, and a synthesized pose, we propose a novel model for synthesizing a human try-on image with the target clothing in the best fitting pose. The qualitative and quantitative experiments both indicate that the proposed TF-TIS outperforms the state-of-the-art methods, especially for difficult cases.
\end{abstract}


%
\IEEEpeerreviewmaketitle

\section{Introduction}
%
%
%
%


 

\IEEEPARstart{S}HOPPING in the brick-and-mortar stores takes considerable time to purchase one satisfactory item of clothing because it usually requires entering a store to try on several clothing candidates. In contrast, shopping online is expected to be a much faster purchase journey because the process of finding the products with relevant items is facilitated by the online searching and recommendation technology. However, although the usage rate of online shopping is rapidly increasing, it is still overshadowed by the brick-and-mortar stores because e-commerce platforms cannot provide sufficient information for
the consumers. Among many promising approaches~\cite{Viton2017,BeautyGlow2019,fashion1,fashion2,fashion3,fashion4,DressAttention2019,DressBest2018} bridging the gap between online and offline shopping, virtual try-on is regarded as the key technology for the online fashion industry to burgeon, as well as a feasible work to bridge the gap between online and offline shopping.

\begin{figure}[t]
\includegraphics[width=8cm]{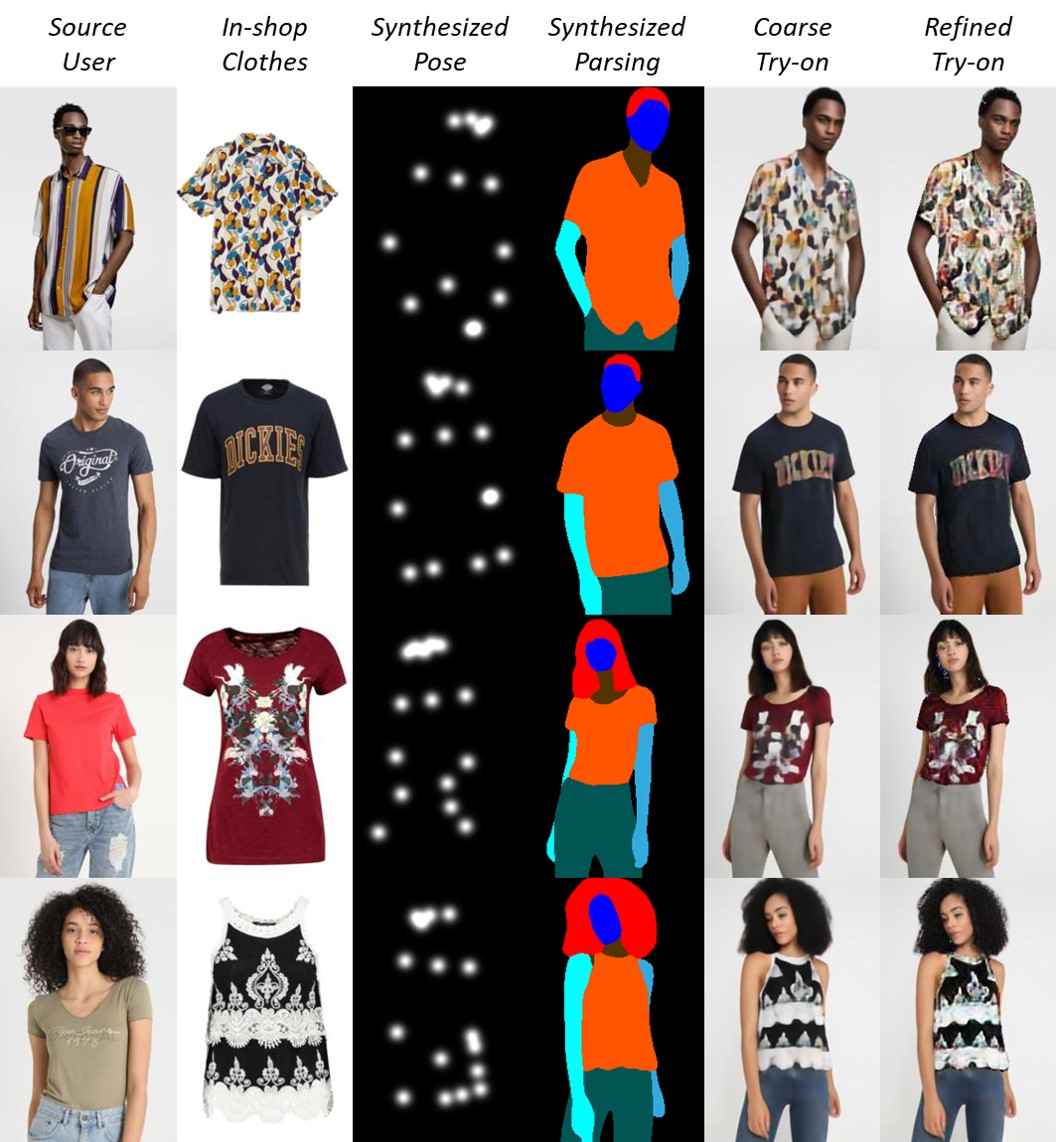}
\centering
    \caption{Examples of our template-free try-on image synthesis (TF-TIS) network, which takes only the in-shop clothing image and user image as input without a defined pose. Our goal is to generate a realistic try-on image according to the synthesized pose from the referential clothing image, which can reduce the cost of hiring photographers. No other work has achieved this.}
\label{fig:7}
\end{figure}

To realize virtual try-on services, a recent line of studies has used clothing warping to transform the in-shop clothing and then paste the warped clothing on user images~\cite{CPVTON2018,Viton2017}, which preserves the details of clothes, including patterns and decorative designs. Nonetheless, the quality of the results significantly decreases when an occlusion (e.g., the user's arm is in front of the chest, obscuring the garment in the source image) or a dramatic pose change (e.g., the limbs are from wide-opened to crossed) occurs. To solve these challenging cases, our previous work~\cite{FashionOn} introduced a semantic-guided method, which uses the semantic parsing to learn the relationship between different poses. However, the clothing part of the virtual try-on results still has some artifacts (i.e., missing details, such as small buttons and local inconsistency, such as distorted plaid), which are important for a try-on service. Moreover, although the state-of-the-art virtual try-on applications~\cite{VirtuallyTryOn, FitMe} have demonstrated the try-on results in arbitrary poses (including our previous work~\cite{FashionOn}), they require users to assign the target poses instead of directly recommending the suitable poses based on the clothing style. Therefore, to create a convenient and practical virtual try-on service, a virtual try-on application that automatically synthesizes a suitable pose corresponding to the target clothing is desirable.

Based on the above observations, in this paper, we propose a novel virtual try-on network, namely, the template-free try-on image synthesis (TF-TIS) framework, for synthesizing high-quality try-on images with automatically synthesized poses. In addition, Fig.~\ref{fig:7} illustrates examples of the template-free virtual try-on. Given a source user image and an in-shop clothing, the goal is to first synthesize the target pose automatically, which is further leveraged to generate the try-on image. Fig. \ref{fig:arc1} presents the TF-TIS framework comprising four modules: 1) 
\textit{cloth2pose}, which synthesizes a suitable pose from the in-shop clothing (Column 3 in Fig. \ref{fig:7}), 2) the \textit{pose-guided parsing translator}, which translates the source pose to semantic segmentation according to the synthesized poses (Column 4 in Fig. \ref{fig:7}), 3) \textit{segmentation region coloring}, which renders the clothing and human information on the semantic segmentation (Column 5 in Fig. \ref{fig:7}), and 4) \textit{salient region refinement}, which polishes the important regions, such as faces and logos (the last column in Fig. \ref{fig:7}).

Specifically, given an in-shop clothing for try-on, we first aim to synthesize a suitable corresponding pose represented as keypoints\footnote{The poses are specified by keypoints, which contain 18 keypoints and each keypoint represents one human body joint.}. One of the basic approaches is to use the images of mannequins wearing corresponding in-shop clothes as the target poses. However, some in-shop clothes may not have corresponding images. Another approach is to cluster the in-shop clothes first and assign the most frequent poses in the cluster as the target pose. Nevertheless, such an approach highly depends on the clustering results, whereas rare/unseen clothes may not find the appropriate poses. Therefore, we propose a novel cloth2pose network to directly learn the relationship between the in-shop garment and the target pose, which leverages the deep features from the pretrained model and then uses the regressor to fit the joint map (i.e., keypoints). To the best of our knowledge, this is the first work to generate suitable poses for corresponding in-shop clothes.

Afterward, given a source user image and a synthesized pose, the goal is to synthesize a realistic try-on image. An intuitive method to tackle difficult cases of the body occlusion or the dramatic pose transfer is offering the body parsing information to the current try-on networks. However, this method is not compatible with existing try-on models because most of the previous try-on works have focused on directly warping the clothing item and pasting the warped clothing onto the users.
Therefore, the \textit{pose-guided parsing translator} is proposed by constructing a deep convolutional network to transform a pose into a semantic segmentation form to guide the learning of the next stage. Semantic segmentation plays a critical role in solving difficult cases. For example, limb parsing provides information for solving dramatic pose changes, whereas limb and clothing parsing offer clues addressing body occlusion issues.

Moreover, to present realistic try-on images to users, we color the transformed semantic segmentation with the appearance of the a human and clothes by using a conditional generative adversarial network (CGAN) in \textit{segmentation region coloring}. Finally, \textit{salient region refinement} focuses on two salient regions for try-on services (i.e., face and clothing) and improves these regions with details to achieve better virtual try-on images. For clothing refinement, We constructed a detail-retaining network, which adopts two encoders to extract relatively important features and global and local discriminators to retain the consistency of images, especially clothing. 

Our previous work is called FashionOn\cite{FashionOn}. We have made several changes in this work, and the contributions are summarized as follows.

\begin{itemize}

    \item We designed a new pose synthesis framework, which directly learns the relationship between in-shop clothes and try-on poses to synthesize a suitable try-on pose. The automatically synthesized poses can facilitate a user-friendly platform without the extra effort of uploading a target pose and exhibit better virtual try-on results to attract customers. To the best of our knowledge, TF-TIS is the first virtual try-on network to provide a suitable pose for the corresponding clothing image.
    
    \item We redesigned the clothing refinement generator (Section III-D-2) composed of two distinct encoders due to the unsatisfying results caused by the same parameters of the encoder for the two input features of in-shop clothing and warped coarse clothing. One is to encode the warped coarse clothing and we integrate it with the very detailed features of the in-shop clothing extracted from the other encoder. In addition, we adopted a UNet-like architecture to avoid losing the warped coarse clothing information, such as shape and color.
    
    \item To enhance the consistency of the generated image, which improves the quality of pictures, we proposed global and local discriminators for our ClothingGAN (Section III-D-2). With the local discriminator, the generator is forced to synthesize the image with more natural details. Using the global discriminator, the generator is forced to generate a realistic picture.

\end{itemize}


\section{Related work}
\subsection{Virtual Try-on} 
Existing virtual try-on approaches can be roughly categorized into 3D-based methods (e.g., 3D body shape) and 2D-based methods (e.g., clothing warping). We first introduce these two approaches and then compare them with TF-TIS.
 
\subsubsection{3D-based Try-on}
To generate more realistic results, numerous approaches~\cite{ClothCap, LearningSSS, GarNet} have used users' 3D body shape measurements and 4D sequence (e.g., video) to offer more information.
For example, with high-resolution videos, Pons-Moll \textit{et al.} \cite{ClothCap} first captured the geometry of clothing on a body to obtain a rough body meshes and then aligned the defined clothing templates to garments of the input scans again to generate more realistic and body-fitting clothes.

Given the high cost of physics-based simulation to accurately drape a 3D garment on a 3D body, Gundogdu \textit{et al.} \cite{GarNet} implemented 3D cloth draping using neural networks. Specifically, they used a PointNet-like model to derive the user information and encoded the garment meshes to obtain the point-wise, patch-wise, and global features for the fitted garment estimation.

In summary, although 3D-based approaches can produce try-on videos, the collection of measurement data can be costly, requiring extensive manual labeling or expensive equipment. Therefore, many scholars have resorted to using rich 2D images, which can be easily found online, to achieve the virtual try-on task. Moreover, the proposed TF-TIS only requires a source image and an in-shop clothing to a synthesize try-on image with the suitable pose.

\subsubsection{2D-based Try-on} To synthesize the try-on images, it is necessary to transform the in-shop clothing to fit users' poses. 
Therefore, spline-based approaches are introduced to achieve this task. Among them, thin plate spline (TPS)~\cite{TPS} has been widely adopted and predominates in the nonrigid transfer of images instead of direct generation using neural networks.
For example, Han \textit{et al.} \cite{Viton2017} presented an image-based virtual try-on network (VITON) that warps in-shop clothes through TPS and cascades a refinement network to generate the warped-clothing details with the coarse-grained person image. However, some details are still missed due to the refinement network~\cite{Viton2017}.

To correct deficiencies in \cite{Viton2017}, Wang \textit{et al.} \cite{CPVTON2018} constructed a two-stage framework, CP-VTON, combining the generated person with the warped clothes through a generated composition mask without adopting the refinement network. 
Moreover, Zheng \textit{et al.} advanced CP-VTON \cite{CPVTON2018} and proposed Virtually Trying on New Clothing with Arbitrary Poses (VTNCAP)\cite{VirtuallyTryOn} by adopting a bidirectional GAN and an attention mechanism, which take the place of the generated composition mask in CP-VTON, to focus more on the clothing region. 

Nevertheless, they still neglected that the facial region is also an important factor to determine the quality of the virtual try-on task and cannot preserve the detailed clothing information (e.g., pleats and shadows) to follow the human poses.
In contrast, TF-TIS was developed as a semantic segmentation-based method that avoids these issues. In addition, TF-TIS preserves the comprehensive details of the in-shop clothes (e.g., patterns and texture) and the realistic human appearance (e.g., hair color and facial features) in accordance with human poses and different body shapes. The visual comparison between TF-TIS and the other mentioned methods is illustrated in Fig.~\ref{fig:result1}.

\begin{figure*}[t]
  \includegraphics[width=\textwidth]{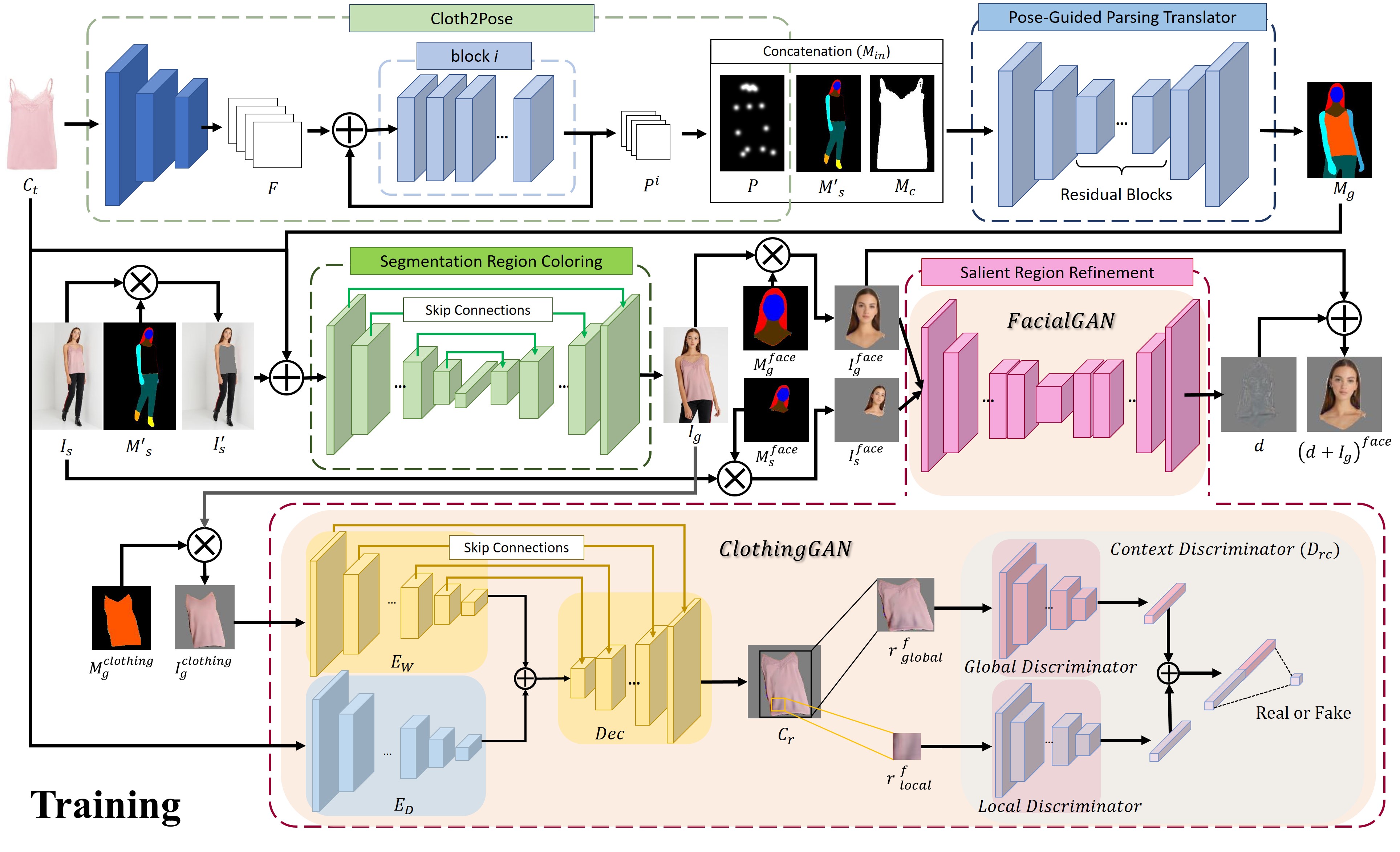}
  \centering
    \caption{\textbf{Training overview}. Stage I (\textit{cloth2pose}) exploits the correlation of clothes and poses and synthesizes a pose from the in-shop clothes via sequential convolution blocks. Stage II (\textit{pose-guided parsing translator}) transfers the human semantic segmentation to $M_g$ according to $M^{'}_s$, $M_c$, and $P$. Stage III (\textit{segmentation region coloring}) fills clothing information and the user's appearance into the segmentation to synthesize a realistic try-on image $I_g$. Stage IV (\textit{salient region refinement}) consists of two parts: FacialGAN and ClothingGAN. FacialGAN generates high-frequency details as a residual output and directly adds on the facial region of $I_g$. ClothingGAN extracts the fine information from the in-shop clothing image and uses the features for the details of the clothes $C_r$.}
  \label{fig:arc1}
\end{figure*}

\subsection{Pose Transfer}
Research on human pose transfer~\cite{Ma2017PoseGP,Multistage,UnsuperPose,DeformableGAN,UnsuperPIG,DenseIntrinsic,ProgressivePoseAttn} has trended recently, as copious applications are planned in the future.
The process of human pose transfer comprises two stages: pose estimation and image generation. The first stage can be divided into two categories (i.e., keypoints estimation \cite{Openpose,LCR,SinglePose} and human semantic parsing~\cite{HumanSemanticParsing,Gong2018InstanceLevelHP,Beyond}).
For example, Hidalgo \textit{et al.}\cite{SinglePose} trained a single-stage network through multi-task learning to determine the keypoints of the whole body simultaneously. Gong \textit{et al.}\cite{Gong2018InstanceLevelHP} used the part grouping network (PGN) to reformulate instance-level human parsing as two twinned sub-tasks that can be jointly learned and mutually refined.

For the second stage, with the advances of GANs, image generation has received considerable attentions and has been widely adopted \cite{Multistage, UnsuperPIG, ProgressivePoseAttn} to generate realistic images. Among the existing pose transfer research, most constructed novel architectures and successfully transferred the pose of the given human image based on the human joint points.
For instance, Ma \textit{et al.}\cite{Ma2017PoseGP} separated this task into two stages: pose integration, which generates initial but blurry images, and image refinement, which refines images by training a refinement network in an adversarial way. Siarohin \textit{et al.}\cite{DeformableGAN} used geometric affine transformation to mitigate the misalignment problem between different poses.
However, most of the previous works did not extend the application of pose transfer to explore virtual try-on. By infusing pose transfer, virtual try-on services provide consumers with more chances to realize their appearance in trying on new clothes in multiple aspects and induce them to buy clothes. Hence, by converting semantic segmentation, TF-TIS seamlessly integrates virtual try-on with pose transfer to generate multi-view try-on images for customers.

\subsection{Cross-modal Learning}
The association between different fields has been studied and exploited recently~\cite{LocalizeSource, VisuallyIndicatedSounds, AudioVisualSceneAnalysis,DeepCMProjection, cross1, IdentityAware, RecurrentResidualFusion,cross2} (e.g., the cross-modal matching between audio and visual signals~\cite{LocalizeSource, VisuallyIndicatedSounds, AudioVisualSceneAnalysis}, image and text~\cite{DeepCMProjection,IdentityAware,RecurrentResidualFusion}).
Castrej{\'o}n \textit{et al.}~\cite{AlignedCMRepresentations} used the identical network architecture with different weights as encoders to extract low-level features from different modalities (e.g., sketches and natural images) and then inputted them into the shared cross-modal representation network to learn the representation for scenes. Tae-Hyun \textit{et al.}~\cite{Speech2Face} attempted to learn voice-face correlation in a self-supervised manner (i.e., directly capturing the dominant facial traits of the person correlated with the input speech instead of synthesizing the face from the attributes).
Inspired by the cross-modal learning, which can find hidden details from an inconspicuous part of data or align the embedding from one domain to another, we used a similar concept to learn the correlation between clothes and human poses to synthesize the image of the virtual try-on with a suitable pose for the user from the corresponding clothing.


\begin{table}[bt]
  \centering
        \caption{Notation Table}
        \begin{tabular}{c|c} 
        \textbf{Symbols} & \textbf{Definitions} \\
        \hline \hline
        $C_t$ & in-shop clothing image \\
        $M_c$ & in-shop clothing mask \\
        $C_d$ & detailed clothing representation\\
        $C_w$ & warped-clothing representation\\
        $C_r$ & refined clothing \\
        \hline
        $P$ & keypoint tensor (suitable pose) \\
        $F$ & in-shop clothing feature map \\
        \hline
        $I_s$ & source user image \\
        $I^{'}_s$ & source user image without clothes\\
        $I_t$ & target user image \\
        $I_g$ & generated try-on image \\
        \hline
        $M_s$ & source body semantic segmentation \\
        $M^{'}_s$ & source body semantic segmentation without clothing part\\
        $M_g$ & generated body semantic segmentation \\
        $M_t$ & target body semantic segmentation \\
        \hline
        $M^{fg}_i$ & foreground channels of $M_i$, $i \in s, g, t$ \\[3pt]
        $M^{face}_i$ & facial channels of $M_i$, $i \in s, g, t$ \\[3pt]
        $M^{clothing}_i$ & clothing channels of $M_i$, $i \in s, g, t$ \\[3pt]
        $I^{face}_i$ & facial part of $I_i$, $i \in s, g, t$ \\[3pt]
        $I^{clothing}_i$ & clothing part of $I_i$, $i \in s, g, t$ \\[3pt]
        \hline
        $d$ & high-frequency residual face details \\
        $N_{c2p}$ & number of convolution blocks in \textit{cloth2pose} \\
        $\otimes$ & pixel-wise multiplication \\
      \end{tabular}
  \label{tab:table2}
\end{table}

\section{Proposed Method}
As illustrated in Fig.~\ref{fig:arc1}, given an in-shop clothing image $C_t$ and a source user image $I_s$, the goal of TF-TIS is to generate the try-on image $I_g$ with an automatically synthesized suitable pose such that the personal appearance and the clothing texture are retained. To achieve this goal, we developed a four-stage framework in TF-TIS: (I) \textit{cloth2pose}, which derives a suitable pose $P$ based on the in-shop clothing $C_t$ by exploiting the correlation between poses and clothes; (II) the \textit{pose-guided parsing translator}, which transforms the body semantic segmentation $M_s$ into a new one, $M_g$, according to the derived pose; (III) \textit{segmentation region coloring}, which takes $I_s$, $C_t$, and $M_g$ as input and synthesizes a coarse try-on image $I_g$ by rendering the personal appearance and clothing information into the segmentation regions; and (IV) \textit{salient region refinement}, which refines the salient but blurry regions of the try-on result $I_g$, generated from the last stage (i.e, FacialGAN refines facial regions and ClothingGAN refines clothing regions). To clarify the definition of each symbol, we created a table (Table \ref{tab:table2}) to illustrate this clearly.


\subsection{Cloth2pose}
A virtual try-on service usually requires three inputs~\cite{VirtuallyTryOn,CPVTON2018, FitMe}: 1) a user image, 2) an in-shop clothing image, and 3) a target pose. One potential improvement is to automatically generate the target pose according to the in-shop clothing because it reduces the users' efforts. Moreover, a suitable pose better demonstrates the in-shop clothing, which may stimulate consumption. For example, plain T-shirts in a sideways pose can mostly show muscle lines. To synthesize the target pose directly from in-shop clothing, \textit{cloth2pose} uses pairs of in-shop clothes and mannequin photos on the online shopping site for training. Specifically, \textit{cloth2pose} first derives keypoints of mannequin photos by existing models, e.g.,~\cite{Openpose,pose1,pose2}.\footnote{We use OpenPose model~\cite{Openpose}, a 2D pose estimation model pre-trained on large-scale human pose datasets (COCO~\cite{MSCOCO} and MPII~\cite{PoseEstimation2014}) in our experiment. The following keypoints are used: nose, eyes, ears, neck, shoulders, elbows, wrists, hips, knees, and ankles.} Let $x_k$ denote the 2D position of the $k^{th}$ keypoint on the image ($I_t$). Because it is difficult to regress the clothing features to a single point, we converted the keypoint position $x_k$ into the pose map $P_k$ by applying a 2D Gaussian distribution for each keypoint. The values at position $p \in {R}^2$ in $P_k$ are defined as follows:
\begin{small}
\begin{equation}
    P_k(p) = \exp{\left( -\frac{\lVert p - x_k\lVert^2_2}{\sigma^2} \right)},
\end{equation}
\end{small}where $\sigma$ determines the spread of the peak. After constructing the 2D keypoint map for each keypoint, we stack all the 2D keypoint maps together as a keypoint tensor, denoted as $P$.

Afterward, \textit{cloth2pose} extracts features of the in-shop clothes by using the first 10 layers of VGG-19~\cite{VGG19}, denoted as $\phi_0$. Let $C_t$ denote the image of in-shop clothing. The clothing feature map $F$ is obtained as $\phi_0(C_t)$. Here, \textit{cloth2pose} exploits a progressive refinement architecture as illustrated in Fig.~\ref{fig:arc1}. Specifically, at the first block, the network produces a set of keypoint information only from the clothing feature map: $P^1 = \phi_1(F)$, where $\phi_1$ refers to the first convolutional block. For the succeeding convolutional blocks, we employed five convolutional layers with a $7\times7$ kernel and two with a $1\times1$ kernel to generate the keypoint tensor, and each layer is followed by a ReLU. The convolutional block takes the concatenation of $F$ and the prediction from the previous block as input to predict the refined keypoint tensor:
\begin{small}
\begin{equation}
    P^i = \phi_i(F, P^{i-1}), \forall 2 \leq i \leq N_{c2p},
\end{equation}
\end{small}where $\phi_i$ represents the $i^{th}$ convolutional block and $N_{c2p}$ is the number of total convolutional blocks in \textit{cloth2pose}.

An intuitive choice for the loss function is the $L_2$ distance between the keypoint tensors extracted from the pose estimation model ($P$) and that estimated from \textit{cloth2pose} ($P^{N_{c2p}}$), i.e., $\lVert P^{N_{c2p}}-P \lVert^2_2$. However, only using the $L_2$ loss is likely to generate many responses in various locations for one joint. Given this condition, we employed the sparsity constraint to limit the number of candidates. Therefore, if the model predicts several candidates for one keypoint, the nonjoint area is penalized less by L2 loss than by L1 loss. The final loss is as follows:
\begin{equation}
    \mathcal{L}_{c2p} = \sum_{i \in N_{c2p}}\lVert P^i-P \lVert^2_2 \ + \  \lambda\lVert P^{N_{c2p}} \lVert_1,
\end{equation}
where $\lambda=0.00008$ is the hyperparameter for striking a balance between multiple candidates and keypoint vanishing.
If the value of $\lambda$ is too high (e.g., $\lambda = 0.001$), then the output $P^{N_{c2p}}$ is without any candidates. Conversely, if the value is too low (e.g., $\lambda = 0.00001$), then the sparsity constraint becomes ineffective, and the output still has more than one candidate.

\subsection{Pose-guided Parsing Translator}

Showing the corresponding area of each body part explicitly, the human body segmentation are employed to synthesize realistic human images. Accordingly, the goal of the \textit{pose-guided parsing translator} is to translate the source body semantic segmentation $M_{s}$ to the target body semantic segmentation $M_{t}$ according to the target pose $P$. We first used the PGN\cite{Gong2018InstanceLevelHP}, which is pretrained on the Crowd Instance-level Human Parsing dataset, to produce semantic parsing labels. The labels contain 20 categories, including left-hand, top clothes, and face. Afterward, to precisely map each item to the new position according to the pose $P$, we used one-hot encoding to constitute a 20-channel tensor $M \in  {R}^{{20}\times{W}\times{H}}$, where each channel is a binary mask representing one category. 
Due to the unnecessity of the clothing channel of $M_{s}$, we replace it with the original in-shop clothing mask $M_c$. This replacement facilitates offering the in-shop clothing shape to realize the virtual try-on service. 

Adapted from pix2pix\cite{Phillip2017pix2pix}, the \textit{pose-guided parsing translator} consists of two downsampling layers, nine residual blocks, and two upsampling layers. Convolutional layers and highway connections, concatenating the input and the output of the corresponding block, are composed in each residual block. The objective of the translator $G_t$ adopts a CGAN as follows:
\begin{small}
\begin{equation}
    \begin{split}
        \mathcal{L}^{G_t}_{GAN}(G_t, D_t) &= \mathbb{E}_{M_{in}, M_t}[log\textit{D}(M_{in}, M_t)] \\
        &+ \mathbb{E}_{M_{in}}[log(1-\textit{D}(M_{in}, G_t(M_{in}))],
    \end{split}
\end{equation}
\end{small}where $G_t$ minimizes the objective against $D_t$ that maximizes it (i.e., $\arg \min_{G_t} \max_{D_t} \mathcal{L}^{G_t}_{GAN}(G_t, D_t)$) and $M_{in}$ represents the concatenation of $M^{'}_s$, $P$, and $M_c$. 

To accurately differentiate each pixel as the corresponding channel, we integrate a pixel-wise binary cross-entropy loss of the $G_t$, denoted as $\mathcal{L}^{G_t}_{BCE}$, with our CGAN objective, and the discriminator stays the same:
\begin{small}
\begin{equation}
\begin{split}
        \mathcal{L}&^{G_t}_{BCE}(G_t) = \\
        &-\sum_{n_c} M_t\log(G_t(M_{in}))+(1-M_t)\log(1-G_t(M_{in})),
\end{split}
\end{equation}\end{small}where $n_c$ denotes the total number of channels of human parsing masks. In summary, the objective of the \textit{pose-guided parsing translator} is derived as follows:
\begin{equation}
    \arg \min_{G_t} \max_{D_t} \mathcal{L}^{G_t}_{GAN}(G_t, D_t)+ \lambda_{bce}\mathcal{L}^{G_t}_{BCE}(G_t).
\end{equation}

\subsection{Segmentation Region Coloring}
Having obtained the target semantic segmentation from the previous stage, the \textit{segmentation region coloring} aims to synthesize a coarse try-on result by rendering information into the segmentation regions, denoted as $M_g = G_t(M_{in})$.
Given the great success of applying GANs in various image generation tasks, we adopte the architecture of CGAN\cite{cGAN} to synthesize results. Specifically, we propose a coloring generator $G_{c}$ rendering the personal information into the body semantic segmentation $M_g$ according to $I_{s}$ and $C_{t}$ (i.e., the appearance of the source person and in-shop clothing texture). Because it is difficult to derive a significant number of training images, we traine our network to change the source person. To avoid supplying $G_c$ with the clothing information, we remove the clothing information from $I_{s}$. In other words, we take as input 1) the in-shop clothing $C_{t} \in {R}^{{3}\times{W}\times{H}}$, 2) the source person image without clothing information $I^{'}_{s}\in {R}^{{3}\times{W}\times{H}}$, and 3) the target semantic segmentation $M_g \in {R}^{20\times{W}\times{H} }$ for $G_{c}$.

Fig.~\ref{fig:arc1} illustrates the architecture of TF-TIS. We adopted the UNet architecture with highway connections, combining the input and processed information. Highway connections were employed to avoid the vanishing gradient \cite{GradDesDiff}.
Six residual blocks were implemented between the encoder and the decoder of $G_c$. For each residual block, two convolutional layers and ReLU were stacked to integrate $M_g$, $I^{'}_{s}$, and $C_{t}$ from small local regions to broader regions so that the appearance information of $I^{'}_{s}$ and $C_{t}$ can be extracted. 

Because the background information is less important and easily distracts the generator from synthesizing try-on images, we filtered out it to force $G_{c}$ to concentrate on generating the correct human part of the image rather than the whole image. Specifically, the background information of the generation result $I_{g}= G_{c}(C_{t},I^{'}_{s},M_{g})$ is filtered out with $M^{fg}_g$ and so is the ground truth $I_{t}$ with $M^{fg}_t$, where $M^{fg}_g$ and $M^{fg}_t$ represent $M_g$ and $M_{t}$ without the background channel, respectively.
Afterward, a global structural information and other low-frequency features are obtained from calculating the L1 distance function:
   \begin{equation}
       \mathcal{L}^{G_c}_{L1}=\sum_{W}\sum_{H} \left\lVert I_{g} \otimes M^{fg}_g -I_{t} \otimes M^{fg}_t\right\lVert_{1},
   \end{equation}
where $\otimes$ represents the pixel-wise multiplication.
   
For the discriminator, we constructed the coloring discriminator $D_{c}$ against $G_{c}$ to distinguish two pairs: one including $I_t$ and $I_s$, and the other including $I_g$ and $I_s$. With the additional real image $I_s$, $D_c$ impels $G_{c}$ to generate more realistic images. Moreover, because this is a binary classification problem (i.e., the image is real or fake), we employed the binary cross-entropy loss as the GAN loss to compare the generated images:

\begin{equation}
    \mathcal{L}^{G_c}_{GAN} = \mathcal{L}_{BCE}(D_{c}(G_{c}(C_{t},I^{'}_{s}, M_g), I_{s}),1),
\end{equation}
\begin{equation}
\begin{split}
        \mathcal{L}^{D_c}_{GAN} &= \mathcal{L}_{BCE}(D_{c}(G_{c}(C_{t},I^{'}_{s},M_g),I_{s}),0)\\
        &+\mathcal{L}_{BCE}(D_{c}(I_{t},I_{s}),1),
\end{split}
\end{equation}
where $G_{c}$ attempts to deceive $D_{c}$ to recognize the synthesized image as a real image; thus, the goal of $\mathcal{L}_{BCE}$ in $\mathcal{L}^{G_c}_{GAN}$ is equal to $1$. 
In contrast, because $D_c$ must classify the generated or real images correctly, the goals of $\mathcal{L}_{BCE}$ in $\mathcal{L}^{D_c}_{GAN}$ are equal to $0$ and $1$, respectively. In summary, the overall loss function of \textit{segmentation region coloring} is as follows:
\begin{equation}
    \mathcal{L}^{G_c} = \mathcal{L}^{G_c}_{GAN} +\lambda \mathcal{L}^{G_c}_{L1}.
\end{equation}

\subsection{Salient Region Refinement}
Because users care most about the characteristics of products, the performance of the virtual try-on service is highly dependent on the saliency of the synthesized image, for example, users (e.g., facial details or body shape), clothing features (e.g., button or bow tie), and 3D physics (e.g., pleat and shadows). Hence, in the fourth stage, we proposed two networks to refine the facial and clothing regions separately.

\subsubsection{FacialGAN}
Modeling faces and hair is challenging but essential in synthesizing try-on images. To simplify this complicated work, our network generates residual face details instead of the whole face. Precisely, for the facial refinement network $G_{rf}$, we adjusted the model of the \textit{segmentation region coloring} ($G_c$) to the facial refinement task by excluding the fully connected layer to avoid losing input details during compression. To force $G_{rf}$ to concentrate on facial details, $M^{face}_g$ and $M^{face}_s$ were introduced to filter out the facial region from $I_{g}$ and $I_{s}$, respectively, where $M^{face}_g$ denotes the parsing channels representing the head (including the face, neck, and hair).
As such, $G_{rf}$ generates the high-frequency details as the residual output $d=G_{rf}(I^{face}_g, I^{face}_s)$, where $I^{face}_g = I_{g}\otimes M^{face}_g$ and $I^{face}_s = I_{s}\otimes M^{face}_s$. After processing images through $G_{rf}$, the fine-tuned result is obtained by adding $d$ to $I_{g}$.

In addition, inspired by \cite{EnhanceNet,perception1}, the perceptual loss was exploited to produce images that have a similar feature representation even though the pixel-wise accuracy is not high.
Let $(d+I_g)^{face}$ and $I^{face}_t$ denote the regions within $M^{face}_g$ of $(d+I_g)$ and $I_t$, respectively.
In addition to calculating the loss pixel-wise $\left\lVert (d+I_g)^{face} -I^{face}_t \right\lVert_{1}$, we computed the perceptual loss by mapping both $(d+I_g)^{face}$ and $I^{face}_t$ into the perceptual feature space through the different layers ($\phi_{i}$) of the VGG-19 model.
This additional loss allows the model to reconstruct the details and edges better. 

\begin{small}
\begin{equation}
\begin{split}
    \mathcal{L}^{G_{rf}}_{vgg}((d+I_{g})&^{face}, I^{face}_t) \\ 
    &=    \sum_{i} \lambda_{i}\left\lVert \phi_{i}((d+I_{g})^{face}) - \phi_{i}(I^{face}_t)\right\lVert_{1}, 
\end{split}
\end{equation}
\end{small}where $\phi_{i}$ represents the feature map retrieved from the $i^{th}$ layer in the pretrained VGG-19 model\cite{VGG19}. Furthermore, like previous stages, we integrated the GAN loss as follows:
\begin{small}
\begin{equation}
    \mathcal{L}^{G_{rf}}_{GAN} =
    \mathcal{L}_{BCE}(D_{rf}((d+I_g)^{face}, I^{face}_s), 1)
\end{equation}
\end{small}

\begin{small}
\begin{equation}
    \begin{split}
    \mathcal{L}^{D_{rf}}_{GAN} &= \mathcal{L}_{BCE} (D_{rf}(I^{face}_s, (d+I_g)^{face}), 0)\\
        &+\mathcal{L}_{BCE}(D_{rf}(I^{face}_s, I^{face}_t), 1).
    \end{split}
\end{equation}
\end{small}

The overall loss function of FacialGAN is as follows:
\begin{small}
\begin{equation}
\begin{split} 
    \mathcal{L}^{G_{rf}} &=
    \lambda_{f1}\mathcal{L}^{G_{rf}}_{GAN}\\
    &+\lambda_{f2}\mathcal{L}^{G_{rf}}_{vgg}((d+I_g)^{face}, I^{face}_t)\\
    &+\lambda_{f3}\sum_{W}\sum_{H} \left\lVert (d+I_g)^{face} -I^{face}_t \right\lVert_{1}\\
    &+\lambda_{f4}\sum_{W}\sum_{H} \left\lVert (d+I_g) \otimes M^{fg}_g - I_t \otimes M^{fg}_t\right\lVert_{1},
\end{split}
\end{equation}
\end{small}

where $\lambda_i$ denotes the weight of the corresponding loss.

\subsubsection{ClothingGAN}
\label{clothingGAN}
Most state-of-the-art virtual try-on networks \cite{Viton2017, CPVTON2018, M2E, VirtuallyTryOn} preserve detailed clothing information by fusing the prewarped clothes onto the try-on images directly. However, these approaches encounter the problems of limbs occlusion or incorrect warping patterns of clothes. To solve these problems, in our previous work (FashionOn)\cite{FashionOn}, we implemented the virtual try-on framework by 1) transforming the human pose into the semantic segmentation form through $G_t$, 2) coloring the clothing textures and human appearance through $G_c$, and 3) processing images through refinement networks. 

Although FashionOn fills in most clothing information back, some tiny but important details (e.g., neckline or button) are missing and the generated images are not sufficient realistic. Hence, we modify the previous clothing refinement generator and construct a new one ($G_{rc}$) to retrieve clothing features directly from the in-shop clothing $C_{t}$ and render them into the clothing region of $I_{g}$. Inputting the concatenation of the in-shop clothing and warped clothing into the Clothing UNet in our previous work~\cite{FashionOn} improved the details, but the generated clothing region still lacks fined details, such as the neckline and buttons. The unsatisfactory results are caused by the same parameters of the encoder for the two input features of in-shop clothing and warped clothing. Moreover, the subtle difference in the details is neglected by the discriminator.

Based on these observations, the proposed ClothingGAN $G_{rc}$ contains four parts: (a) detail encoder ($E_D$), (b) warped-clothing encoder ($E_W$), (c) decoder ($Dec$), and (d) context discriminator ($D_{rc}$). The generator exploits detailed information on in-shop clothing and warped clothing obtained from $E_D$ and $E_W$, respectively, which are then input into $Dec$ to generate an image of refined clothing. Next, $D_{rc}$, which consists of the local and the global discriminators, differentiates whether the refined clothing is real or fake by comparing the local and global consistency with real images.

\textbf{Detail Encoder ($E_D$).}
The objective of $E_D$ is to learn the detailed and neglected information (i.e., missing information in the previous stage) from an in-shop clothing image ($C_t$). To extract detailed visual features, we use seven convolutional layers followed by an instance normalization (IN) layer\cite{improvedtexture} together with LeakyReLU\cite{LeakyReLU} as the activation function, which is more than $E_W$ because detailed information is required from the original in-shop clothing, such as texture and logos. After training, $E_D$ can generate a detailed clothing representation, denoted as $C_d = E_D(C_t)$, which is further employed by the decoder to complement the details and synthesize the refined clothing. 

\textbf{Warped-Clothing Encoder ($E_W$).}
As depicted in Fig.~\ref{fig:arc1}, we use the UNet architecture to encode $I^{clothing}_g = I_{g} \otimes M^{clothing}_g$, where $M^{clothing}_g\in {R}^{{W}\times{H}}$ is the clothing part of $M_{g}$. The encoder includes five downsampling convolutional layers with kernel=5, and each layer is followed by an IN layer with LeakyReLU. Each layer of the UNet encoder is connected to the corresponding layer of the UNet decoder through highway connections to produce high-level features. Finally, we obtain the warped-clothing representation $C_w = E_w(I^{clothing}_g)$. In the following section, we present how the outputs of $E_D$ and $E_W$ have been further employed in the decoder network.

\textbf{Decoder ($Dec$).}
To generate refined clothing via the decoder, we concatenate the encoded features $C_d$ and $C_w$ obtained from $E_D$ and $E_W$, respectively, as input. From layer to layer in the decoder, we first derive the features obtained from the previous layer and the precomputed feature maps at $E_W$ connected through a highway connection. Next, we upsample the feature map with the $2\times2$ bicubic operation. After upsampling, a $3\times3$ convolutional and ReLU operation are applied. Using the highway connections with $E_W$ allows the network to align the detailed clothing features with the warped-clothing features obtained by the UNet encoder ($E_W$). In other words, the generator can be written as follows:
\begin{small}
\begin{equation} 
    \label{eq:15}
    \begin{split}
        C_r &= G_{rc}(C_{t}, I^{clothing}_g) \\ 
        &= Dec(E_D(C_{t}), E_W(I^{clothing}_g)).
    \end{split}
\end{equation}
\end{small}

To bridge the difference between the refined clothing $C_{r}$ and the target clothing region $I^{clothing}_t = I_{t} \otimes M^{clothing}_t$, where $M^{clothing}_t$ represents the clothing channel of $M_{t}$, we introduced the L1 loss ($ \mathcal{L}^{G_{rc}}_{L_1}$) and the perceptual loss ($\mathcal{L}^{G_{rc}}_{vgg}$) to refine the clothing as follows:

\begin{small}
\begin{equation}
    \label{eq:12}
    \mathcal{L}^{G_{rc}}_{L_{1}}(C_{r},I^{clothing}_t)=\sum_{W}\sum_{H} \left\lVert C_{r}-I^{clothing}_t\right\lVert_{1},
\end{equation}
\end{small}

\begin{small}
\begin{equation}
    \label{eq:11}
    \mathcal{L}^{G_{rc}}_{vgg}(C_{r},I^{clothing}_t)=\sum_{i=1}^{5} \lambda_{i}\left\lVert 
    \phi_{i}(C_{r})-\phi_{i}(I^{clothing}_t)\right\lVert_{1},
\end{equation}
\end{small}where $\phi_{i}(C)$ represents the feature map of the clothing $C$ of the $i^{th}$ layer in the VGG-19 model \cite{VGG19}. By exploiting the L1 loss instead of L2 loss here, we address the problems of blurry generated images. To further avoid the misalignment, the refined clothing $C_{r}$ is integrated into $I_{g}$, where the clothing region is removed, to synthesize a refined human $I_{rg} = C_{r} \otimes M^{clothing}_g + I_{g} \otimes (1-M^{clothing}_g)$. The parsing mask $M^{clothing}_g$ is used to select the clothing region, which facilitates the process of excluding limbs in front of the clothing when fusing the clothing. The loss for the refined clothing try-on is defined as follows:
\begin{small}
\begin{equation}
    \label{eq:13}
    \mathcal{L}^{G_{rc}}_{fullbody}(I_{rg},I_{t})=\sum_{W}\sum_{H} \left\lVert I_{rg}-I_{t}\right\lVert_{1}.
\end{equation}
\end{small}

\textbf{Context Discriminator.}
To make the refined clothing more realistic, we also employed the GAN loss $ \mathcal{L}^{G_{rc}}_{GAN}$ by adopting the context discriminator comprising the global and the local discriminators that classify the refined clothing as real or fake by comparing the local and the global consistency with real images. Both discriminators are based on a convolutional network that compresses the images into small feature tensors. A fully connected layer is applied to the concatenation of the output feature tensors and predicts a constant value between $1$ and $0$, which represents the probability that the refined clothing is real.

The global discriminator takes as input the image in which we create a bounding box of the clothing part from the result and resizes it, using bilinear interpolation, to $128\times128$. It consists of five two-stride convolutional layers with kernel=5 and a fully connected layer that outputs a 1024-dimensional vector. The local discriminator follows a similar pattern, except the last two single-stride convolutional layers have a kernel=3 and an input size of $64\times64$. The input of the local discriminator is generated by randomly sampling $16\times16$ from the bounding box and resizing it to $64\times64$.

After deriving the outputs from the global and the local discriminators, we build a fully connected layer, followed by a sigmoid function to process the concatenation of two vectors (a 2048-dimensional vector). The output value ranges from 0 to 1, representing the probability that the refined clothing is real, rather than generated. The GAN loss is defined as follows:
\begin{small}
\begin{equation}
    \mathcal{L}^{G_{rc}}_{GAN}= \mathcal{L}_{BCE}(D_{rc}(r^f_{local}, r^f_{global}), 1),
\end{equation}
\end{small}
\begin{small}
\begin{equation}
  \begin{split}
    \mathcal{L}^{D_{rc}}_{GAN}&= \mathcal{L}_{BCE}(D_{rc}(r^f_{local}, r^f_{global}), 0) \\
    & +\mathcal{L}_{BCE}(D_{rc}(r^t_{local}, r^t_{global}), 1),
  \end{split}
\end{equation}
\end{small}where $r$ is the resized result, the subscript $global$ or $local$ denoted the whole or sub-sampled result, respectively, and the superscript $t$ or $f$ means that result is true or fake (generated), respectively. The overall loss function of the ClothingGAN is defined as follows:
\begin{small}
\begin{equation}
    \label{eq:14}
    \mathcal{L}^{G_{rc}} = 
    \mathcal\lambda_{c1} \mathcal{L}^{G_{rc}}_{vgg} +
    \lambda_{c2} \mathcal{L}^{G_{rc}}_{L_{1}} +
    \lambda_{c3} \mathcal{L}^{G_{rc}}_{fullbody} +
    \lambda_{c4}\mathcal{L}^{G_{rc}}_{GAN},
\end{equation}
\end{small}where $\lambda_{ci}\ (i=1,2,3,4)$ denotes the weight of the corresponding loss.

\section{Experiments}
The datasets and implementation are detailed here. Afterward, we conduct qualitative and quantitative experiments with the state-of-the-art method and our previous work FashionOn~\cite{FashionOn} to demonstrate the effectiveness of TF-TIS.

\begin{figure*}[t]
\includegraphics[width=\textwidth]{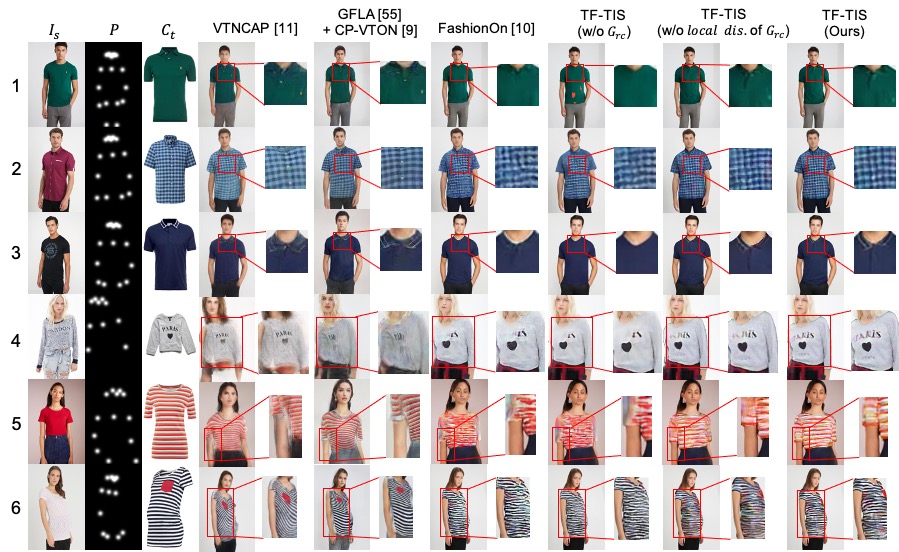}
\centering
    \caption{\textbf{Visual detail comparison.} To compare the details of generated images between different models, we excluded the \textit{cloth2pose} module from our network. The leftmost three columns are the input, and the rest of the columns are the output of different models and the local enlargement of them. Our TF-TIS has the best performance regarding details, such as the neckline of polo shirts and clothing pattern, and retains global and local consistency.}
\label{fig:result1}
\end{figure*}

\subsection{Dataset}
To train and evaluate the proposed TF-TIS, a dataset containing two different poses and one clothing image for each person is required. Still, most of the existing datasets provide either only one pose for each person with the corresponding clothing image~\cite{Viton2017,CPVTON2018}  or multiple poses for each person but without clothing images\cite{DeepFashion2016}. Therefore, we collected a new large-scale dataset containing $10,895$ in-shop clothes with the corresponding images of mannequins wearing in-shop clothes in two different poses.\footnote{Please refer to the images in \url{https://github.com/fashion-on/FashionOn.github.io}.}
In addition, the DeepFashion dataset \cite{DeepFashion2016}, with a size of $288\times192$, is also adopted to broaden the diversity of the data. After removing the incomplete image pairs and wrapping one in-shop clothing and two human images into each triplet, $11,283$ triplets were created. Finally, we randomly split the dataset into the training set and the testing set with $9,590$ and $1,693$ triplets, respectively.

\begin{figure}[bth]
  \includegraphics[width=8cm]{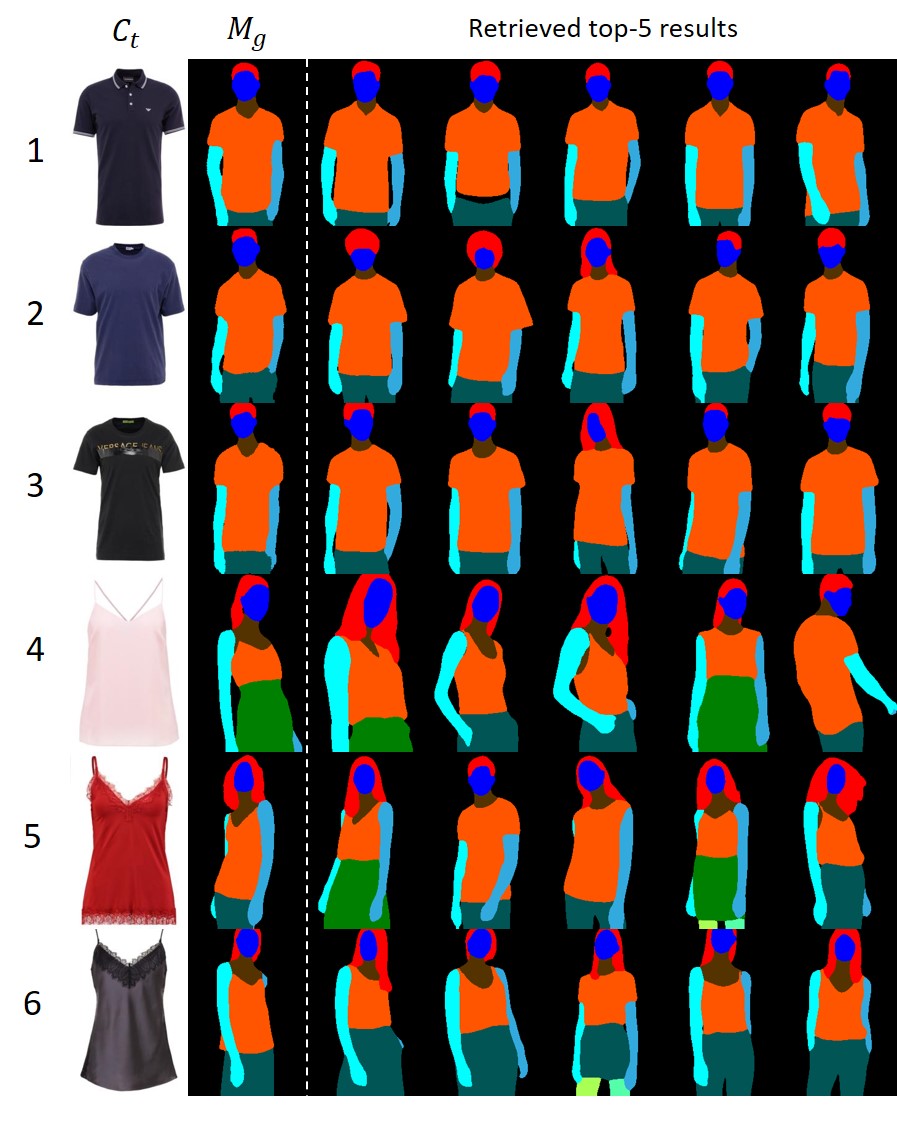}
  \centering
    \caption{\textbf{Pose retrieval examples of \textit{cloth2pose}.} We queried our training dataset by comparing the features extracted via \textit{cloth2pose} to all clothing features in the dataset. For each query, we present the top five retrieved samples. To focus on the pose information, we eliminate the human information, such as skin or hair color. The leftmost two columns are input clothes and the translated parsing, generated via Stage II from the derived pose. Although some examples are not like the query, it still shows that we could easily find results visually close to the query.}
  \label{fig:retrieval}
\end{figure}

\subsection{Implementation Details}

\textbf{Cloth2pose.} We initialize the first 10 layers with that of the VGG-19~\cite{VGG19} and fine-tune them to generate a set of clothing feature maps $F$ from the information on the in-shop clothing. For the following convolutional blocks, each contains five convolutional layers with a $7\times7$ kernel and two with a $1\times1$ kernel. Each layer is followed by a ReLU. In this stage, we apply $N_{c2p} = 4$ to the number of the convolutional blocks.

\textbf{Pose-guided Parsing Translator.} Based on the framework of ResNet, we implement two downsampling layers, nine residual blocks, and followed by two upsampling layers. Specifically, we construct two single-stride convolutional layers with a $3\times3$ kernel and one highway connection, combining the input and the output of each corresponding residual block.

\textbf{Segmentation Region Coloring.} The architecture is composed of the encoder and decoder with six residual blocks between them. Except for the last residual block and one fully connected layer, each block contains two single-stride convolutional layers with a $3\times3$ kernel, one downsampling two-stride convolutional layer with a $3\times3$ kernel. The number of filters of all convolutional layers linearly increases and decreases, respectively, for the encoder and decoder.

\textbf{Salient Region Refinement.} The generator of FacialGAN ($G_{rf}$) is similar to $G_c$ but without the fully connected layer. In addition, $G_{rf}$ has four residual blocks containing two convolutional layers and one downsampling convolutional layer. For ClothingGAN, the generator ($G_{rc}$) comprises two different encoders and one decoder. The detail encoder ($E_D$) consists of four downsampling convolutional layers and three convolutional layers, and the warped-clothing encoder ($E_W$) consists of four downsampling convolutional layers and one convolutional layer. All downsampling convolutional layers have a $4\times4$ kernel and a $2\times2$ stride, and other convolutional layers have a $3\times3$ kernel and a $1\times1$ stride. Both kinds of convolutional layers are followed by the IN layer and LeakyReLU. The decoder ($Dec$) consists of five $3\times3$ convolutional layers, and each layer is followed by one upsampling layer, one IN layer, and one ReLU. 

For the context discriminator ($D_{rc}$), we adopt two discriminators: (1) the global discriminator, which consists of four downsampling convolutional layers and outputs a 1024-dimensional vector representing the global consistency, and (2) the local discriminator, which consists of three downsampling convolutional layers and outputs a 1024-dimensional vector representing the local consistency. A fully connected layer and sigmoid function are applied to the concatenation of the two vectors to differentiate whether the image is real or generated. 

We used Adam \cite{Adam} with $\beta_1 = 0.5$ and $\beta_2 = 0.999$ as the optimizer for all stages. The learning rates of the \textit{pose-guided parsing translator} and the other stages are 2e-4 and 2e-5, respectively. 

\subsection{Qualitative Results}

\begin{figure}[bt]
  \includegraphics[width=8.5cm]{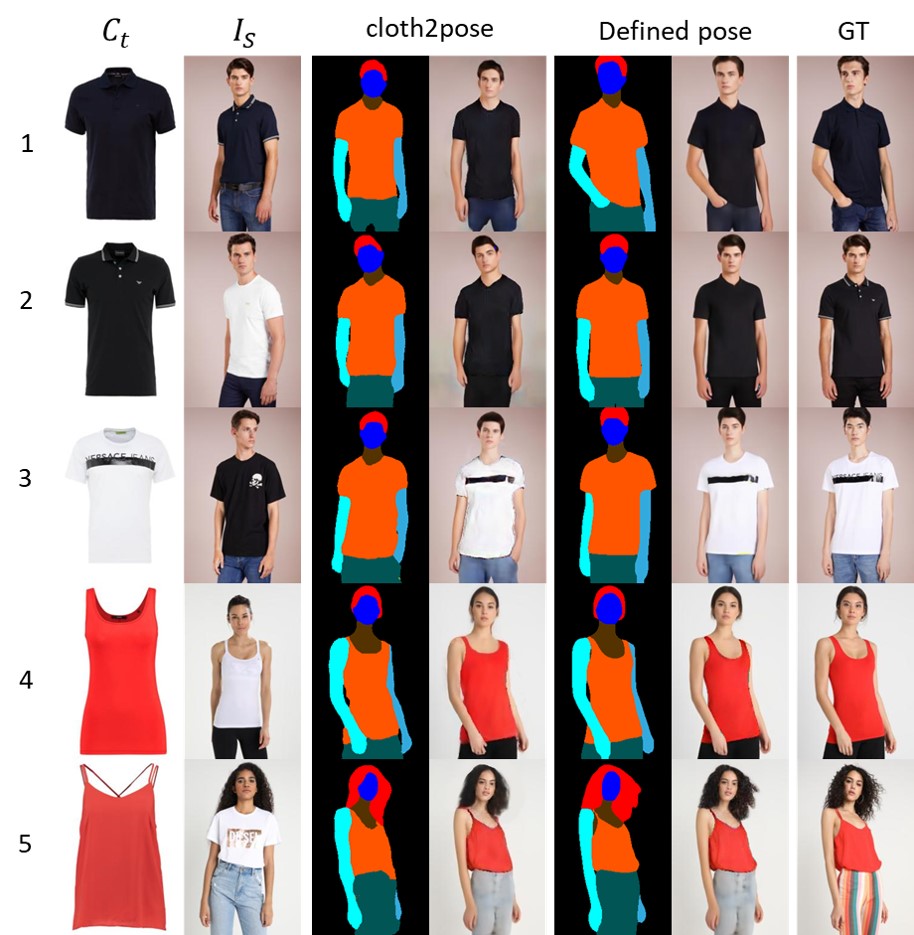}
  \centering
    \caption{\textbf{Qualitative results sampled from our testing dataset.} For every example (six images as a group) we show from left to right is: the input clothing, the generated segmentation image with the synthesized pose from TF-TIS, the try-on result with the synthesized pose, the generated segmentation image with the defined pose in our dataset, the try-on result with the defined pose, and the real try-on image.}
  \label{fig:result2}
\end{figure}

\begin{figure*}[bth!]
    \includegraphics[width=0.93\textwidth]{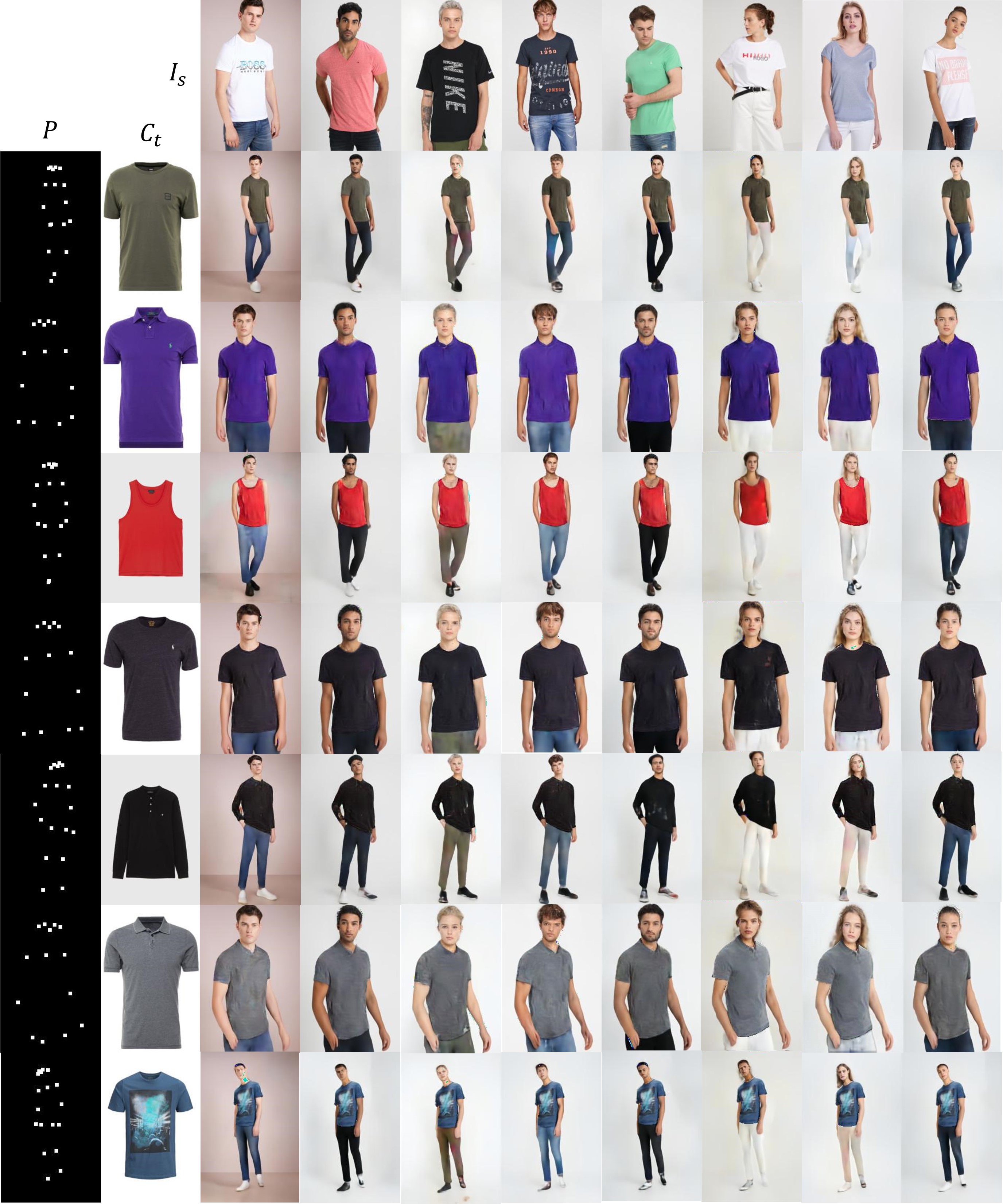}
    \centering
    \caption{Qualitative results sampled from the testing dataset.}
    \label{fig:mmbig}
\end{figure*}

Several try-on results are depicted in Fig~\ref{fig:result1}, \ref{fig:result2}, \ref{fig:AdaIN}, and \ref{fig:mmbig}.
\subsubsection{Evaluation of Virtual Try-on}
As Fig.~\ref{fig:result1} reveals, we compare TF-TIS with the state-of-the-art clothing warping-based method (VTNCAP~\cite{VirtuallyTryOn}) and our previous work (FashionOn~\cite{FashionOn}), which adopts a coarse-to-fine strategy. In addition, because CP-VTON does not include the pose transfer, we combine the state-of-the-art pose transfer method GFLA~\cite{gfla} with CP-VTON~\cite{CPVTON2018} as an additional baseline (GFLA+CP-VTON). The results indicate that all methods accomplish the task of virtual try-on with arbitrary poses. However, the results of VTNCAP and GFLA+CP-VTON contain some artifacts, while the results of FashionOn lose some details and local consistency. Several cases are worth mentioning and listed below.

\noindent \textbf{Neglecting Tiny but Essential Details.}
Fig.~\ref{fig:result1} illustrates that the ClothingGAN ($G_{rc}$) does generate detailed information. From the left to the right, the results are from the state-of-the-art works (VTNCAP, GFLA+CP-VTON, and FashionOn) and ablation studies for $G_{rc}$ in TF-TIS (TF-TIS without $G_{rc}$, TF-TIS without the local discriminator, and TF-TIS). The approaches without $G_{rc}$ (two encoders), including VTNCAP, GFLA+CP-VTON, and FashionOn, fail at the erroneous neckline and the small button, as revealed in Rows 1 and 3. The neckline and the small button on the clothing image by FashionOn are neglected because FashionOn uses only one encoder to extract the information of the concatenation of the in-shop clothing and warped clothing, which degrades the focus of both images. In contrast, the local discriminator of TF-TIS discerns tiny clothing details and the global discriminator is applied to retain the consistency of the entire image. As a result, TF-TIS generates the neckline and small button based on more comprehensive information of the warping clothing, which generates an appearance that is closer to the in-shop clothing images.

\noindent\textbf{Wrong Warping Pattern.} As depicted in Row 2 in Fig.~\ref{fig:result1}, FashionOn and TF-TIS successfully resolve the wrong warping pattern problems of VTNCAP. Because warping clothes through TPS \cite{TPS} only considers the deformation of clothes in two dimensions, the warped clothes are unrealistic. Although in Rows 1 and 2 GFLA+CP-VTON preserves the neckline and the button and generates smooth plaid, GFLA+CP-VTON misses the shade and makes the clothes an average color in Row 4. In Row 6, GFLA+CP-VTON mistakes the red pocket as being on the right side. In contrast, we predict the warped-clothing mask based on the in-shop clothing mask and the warped body segmentation, which consider the correlation between body parts. Moreover, the proposed TF-TIS retains the consistency of clothes, such as the pattern shape, which makes the plaid shirts more realistic because we adopt global and local discriminators to discern the clothing details and to retain consistency.

\noindent\textbf{Average Face.} The VTNCAP often synthesizes an average face as depicted in the fourth column of Fig.~\ref{fig:result1}, because it simply uses the whole body as a mask and renders the human information into it. In contrast, we treat human parsing using 18 channels and render the information for each body part into the corresponding region, which is more specific for every part. Additionally, our works employs the FacialGAN to refine the facial part, making it more distinctive, instead of synthesizing the average faces.

\noindent\textbf{Clothing Color Degradation.} In the second, fourth, fifth and sixth rows in Fig.~\ref{fig:result1}, the clothing color of the results derived by VTNCAP changes from the color of the in-shop clothing. In contrast, FashionOn and TF-TIS successfully preserve the color of the in-shop clothing, which is important in virtual try-on services.

\begin{figure}[bth]
  \includegraphics[width=8.5cm]{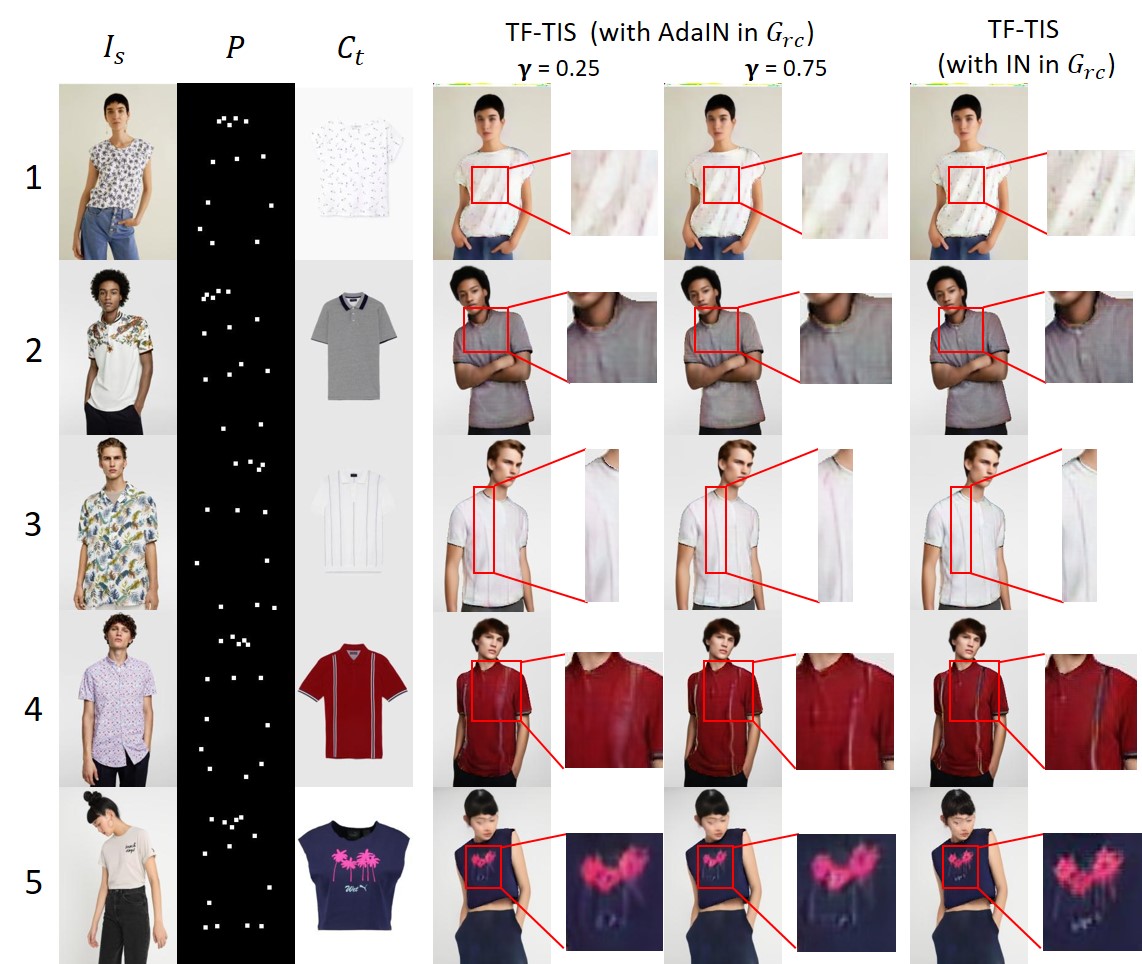}
  \centering
    \caption{Visual comparison of AdaIN~\cite{AdaIN} and IN~\cite{improvedtexture} for ClothingGAN.}
  \label{fig:AdaIN}
\end{figure}

\noindent\textbf{Human Limbs Occlusion.} Rows 5 and 6 in Fig.~\ref{fig:result1} reveal that the proposed TF-TIS can solve the human limbs occlusion problems in VTNCAP. Rather than simply warping it through TPS, we simultaneously warp the clothing and the body segmentation, then render the human appearance and the clothing information sequentially. Hence, $G_c$ can easily render the appearance based on all semantic segmentation, preserving the natural correlation between clothes and humans.

\noindent\textbf{Dropping the Detailed Logo.} In Fig.~\ref{fig:result1}, the rightmost two columns are the ablation study for the local discriminator within the context discriminator. Row 4 shows that the local discriminator generates the full logo. The ``PARIS'' logo is evident with almost all five characters, using the local discriminator in the rightmost column. Without the local discriminator, it only generates three characters.

\noindent\textbf{Comparison of AdaIN and IN for $G_{rc}$.} We replace the IN layer in the two encoders of the ClothingGAN with an adaptive instance normalization layer (AdaIN) to evaluate whether AdaIN helps preserve the clothing details in Fig.~\ref{fig:AdaIN}. Equation~\ref{eq:15} for AdaIN becomes the following:
\begin{small}
\begin{equation} 
\begin{split}
        C_r = Dec(&(1-\gamma)E_W(I^{clothing}_g)\\
        &+\gamma AdaIN(E_W(I^{clothing}_g),E_D(C_{t}))),
\end{split}
\end{equation}
\end{small}where $\gamma$ is a hyperparameter for the content-style trade-off. We used $\gamma=0.25$ and 0.75 to evaluate the difference and demonstrated the visual comparison in Fig.~\ref{fig:AdaIN}.

\begin{small}
\begin{equation}
    {AdaIN}(x,y)=\alpha(y){\left( \frac{x-\mu(x)}{\alpha(x)} \right)}+\mu(y),
\end{equation}
\end{small}where $x$ represents the content input, $y$ is the style input, and $\alpha(y)$ denotes the standard deviation of $y$. The AdaIN simply scales the normalized content input with $\alpha(y)$ and shifts it using $\mu(y)$. Fig.~\ref{fig:AdaIN} reveals that AdaIN tends to generate global features for the clothing information and fails to generate robust details. For example, as presented in Row 4, AdaIN fails to synthesize the robust edge of the suspenders. Moreover, as displayed in Row 5, AdaIN tends to generate the blurry flowers. When increasing the hyperparameter $\gamma$ to contain a higher proportion of features from $C_t$, the $G_{rc}$ adopting AdaIN generates more robust but still blurrier results than using IN. 

\subsubsection{Evaluation of cloth2pose}
Because none of the previous research can generate the target poses according to the in-shop clothes, we evaluate the performance of \textit{cloth2pose} by determining whether \textit{cloth2pose} can learn the relationship between the in-shop clothing and try-on pose. Specifically, in the testing phase, given the in-shop clothing, we use \textit{cloth2pose} to derive the synthesized pose and generate the translated parsing (second column in Fig.~\ref{fig:retrieval}). Afterward, we compute the L2 distances between the in-shop clothing feature and all clothing features in the training dataset and retrieve top five try-on poses results with the smallest clothing distance. The in-shop clothing features are extracted by using the first 10 layers of the VGG-19~\cite{VGG19}.

Fig.~\ref{fig:retrieval} presents several examples. The retrieved results reveal that the synthesized poses are very close to some real poses in the top five results (e.g., the fourth sample in Row 2, the first sample in Row 5, and the first sample in Row 6). Moreover, our retrieved examples also demonstrate that different poses should be synthesized in accordance with the in-shop clothing to better present the clothing. For example, T-shirts, like the clothes in Rows 1 to 3, are demonstrated in the front views to show the logo or with one hand in the pocket to show the muscles. However, the camisole tops in Rows 4 to 6, are demonstrated with people standing sideways to show their body shapes, facing the right or left.

Moreover, the qualitative results of our testing dataset are presented in Fig.~\ref{fig:result2}, and indicate that our model can synthesize a better pose to display clothing. For each example, we present the input clothing ($C_t$), the user ($I_s$), the translated human parsing with the synthesized pose via the \textit{cloth2pose} module and the generated image, the human parsing with the defined pose and the generated image, and the ground truth image of the defined pose. Although appearing a little different from the image with the defined pose, the \textit{cloth2pose} results capture the key information about the human, such as the direction they face. Moreover, we synthesize suitable poses for clothes. For instance, 1) in Row 3, we derive the pose in the front view to show the pattern of the clothing and 2) in Row 4 to 5, we synthesize the sideways pose to show the upper arms and shoulders of people. Therefore, our model understands the relation between clothes and poses and can synthesize better poses to present better try-on results, which induces users to buy clothes.

\subsection{Quantitative Results}
Because the structural similarity (SSIM)~\cite{SSIM} and inception score (IS)~\cite{ImprovedGAN} are fairly standard metrics that focus on the overall quality of the generated image instead of the pixel-wise comparison, we calculated them for the reconstruction of the try-on results in our dataset. 
The SSIM measures the similarity by comparing the generated images against the original images in the structural information, whereas IS provides scores to indicate whether the generated results are visually diverse and semantically meaningful.
  
Compared with the other virtual try-on systems (i.e., VTNCAP, CP-VTON, GFLA+CP-VTON, and FashionOn), our method outperforms them in terms of SSIM and IS, as revealed in Table~\ref{tab:table1}.
Moreover, TF-TIS outperforms  VTNCAP and CP-VTON in terms of IS by $18.9\%$ and $8.14\%$, respectively. Additionally, the comparison in term of SSIM indicates that TF-TIS exceeds VTNCAP and CP-VTON by $19.8\%$ and $11.5\%$, respectively. Although TF-TIS only surpasses the results of FashionOn within 1\% in both metrics, the result complements the important details and the local and global consistency that FashionOn lacks, as demonstrated in Fig~\ref{fig:result1}.

\begin{table}[t]
  \centering
  \caption{\textbf{Comparison of the virtual try-on testing dataset.} We randomly sampled 1300 data from the testing dataset.}
  \begin{tabular}{l|c c} 
    \textbf{Method} & \textbf{IS} & \textbf{SSIM} \\
    \hline \hline
    VTNCAP~\cite{VirtuallyTryOn} & 2.5874 $\pm$ 0.0965 & 0.7282 \\
    CP-VTON \cite{CPVTON2018} & 2.8495 $\pm$ 0.0832 & 0.7824 \\
    GFLA \cite{gfla} + CP-VTON \cite{CPVTON2018} & 3.0266 $\pm$ 0.1740 & 0.8070 \\
    FashionOn (w/o refine) & 3.0679 $\pm$ 0.1247 & 0.8689 \\
    FashionOn (w/ refine)~\cite{FashionOn} & 3.0693 $\pm$ 0.1560 & 0.8724\\
    TF-TIS (Ours) & \textbf{3.0777 $\pm$ 0.1143} & \textbf{0.8725} \\
    \hline
    Real Data & 3.2350 $\pm$ 0.1282 & 1 \\
  \end{tabular}
  
    \footnotesize Note: IS: inception score; SSIM: structural similarity. The higher the score, the better the result.
  \label{tab:table1}
\end{table}

\textbf{Runtime.} We evaluated the efficiency of the proposed TF-TIS by separately reporting the running time of the four modules. The results of the runtime were conducted on a NVIDIA 1080-Ti GPU and were averaged with 2000 randomly selected image sets. The runtime of each module is as follows: \textit{cloth2pose} (1.3 ms), \textit{pose-guided parsing translator} (2.6 ms), \textit{segmentation region coloring} (3.1 ms), and \textit{salient region refinement} ($G_{rf}$: 1.9 ms, $G_{rc}$: 2.6 ms). The results indicate that the proposed TF-TIS not only reduces the cost of hiring photographers but also provides a real-time try-on service for fashion e-commerce platforms.

\section{Conclusion and Future work}
In this paper, we present a part-level learning network (TF-TIS) for virtual try-on service with automatically synthesized poses. The previous work requires a user-specified target pose for try-on. In contrast, TF-TIS precisely generates try-on images with the poses synthesized from the clothing characteristics, which better demonstrates the clothes. The experimental results indicate that TF-TIS significantly outperforms the state-of-the-art virtual try-on approaches on various clothing types, is better in term of being lifelike in appearance, and recommends poses that induce customers to buy clothes. Moreover, as shown in the experiments, TF-TIS captures the relation between clothes and poses to synthesize better poses to present users with better try-on results. In addition, by proposing the global and the local discriminators in the clothing refinement network, TF-TIS retains consistency of images and preserves critical human information and clothing characteristics. Therefore, TF-TIS resolves many challenging problems (e.g., generating tiny but essential details and preserving detailed logos). In the future, we plan to extend our approach to learn how different garment sizes deform on a real body in images using transfer training from 3D human model methods.

\section*{Acknowledgments}
This work was supported in part by the Ministry of Science and Technology of Taiwan under Grants MOST-109-2221-E-009-114-MY3, MOST-109-2218-E-009-025, MOST-109-2221-E-009-097, MOST-109-2218-E-009-016, MOST-109-2223-E-009-002-MY3, MOST-109-2218-E-009-025 and MOST-109-2221-E-001-015, in part by the National Natural Science Foundation of China under Grant 61772043, in part by the Fundamental Research Funds for the Central Universities, and in part by the Beijing Natural Science Foundation under Contract 4192025.


%





\ifCLASSOPTIONcaptionsoff
  \newpage
\fi



\bibliographystyle{IEEEtran}
\bibliography{ref}

%



%
\begin{IEEEbiography}[{\includegraphics[width=1in,clip,keepaspectratio]{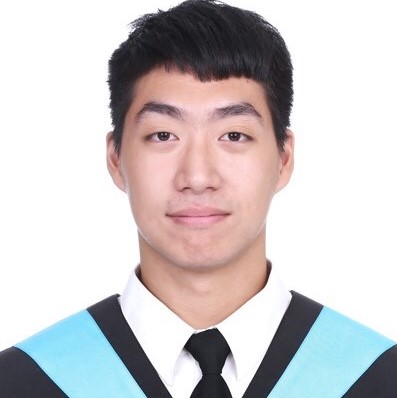}}]{Chien-Lung Chou} received the B.S. degree from the Department of Electrical and Computer Engineering, National Chiao Tung University (NCTU), Hsinchu, Taiwan, in 2019. He is now a master student in the Department of Electrical and Computer Engineering, University of Michigan. His research interest includes artificial intelligence, deep learning, and computer vision.
\end{IEEEbiography}
\begin{IEEEbiography}[{\includegraphics[width=1in,clip,keepaspectratio]{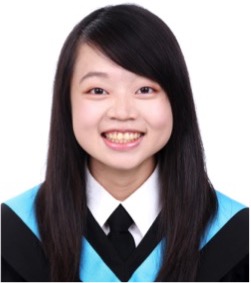}}]{Chieh-Yun Chen} received the B.S. degree from the Deparment of Elecrtophysics, National Chiao Tung University (NCTU), Hsinchu, Taiwan, in 2020. She is now a master student in the Institute of Electronics, NCTU. Her research interests are mainly in artificial intelligence, deep learning, and computer vision.
\end{IEEEbiography}

\begin{IEEEbiography}[{\includegraphics[width=1in,clip,keepaspectratio]{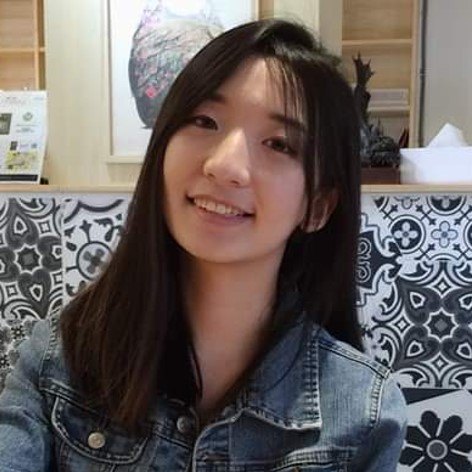}}]{Chia-Wei Hsieh} received the B.S. degree from the Department of Electrical and Computer Engineering, National Chiao Tung University (NCTU), Hsinchu, Taiwan. She is now master student in Electrical and Computer Engineering - Machine Learning and Data Science, University of California, San Diego (UCSD). Her interest includes machine learning and computer vision.
\end{IEEEbiography}

\begin{IEEEbiography}[{\includegraphics[width=1in,height=1.25in,clip,keepaspectratio]{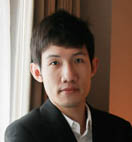}}]{Hong-Han Shuai} received the B.S. degree from the Department of Electrical Engineering, National Taiwan University (NTU), Taipei, Taiwan, R.O.C., in 2007, the M.S. degree in computer science from NTU in 2009, and the Ph.D. degree from Graduate Institute of Communication Engineering, NTU, in 2015. He is now an associate professor in NCTU. His research interests are in the area of multimedia processing, machine learning, social network analysis, and data mining. His works have appeared in top-tier conferences such as MM, CVPR, AAAI, KDD, WWW, ICDM, CIKM and VLDB, and top-tier journals such as TKDE, TMM and JIOT. Moreover, he has served as the PC member for international conferences including MM, AAAI, IJCAI, WWW, and the invited reviewer for journals including TKDE, TMM, JVCI and JIOT.
\end{IEEEbiography}
\begin{IEEEbiography}[{\includegraphics[width=1in,height=1.25in,clip,keepaspectratio]{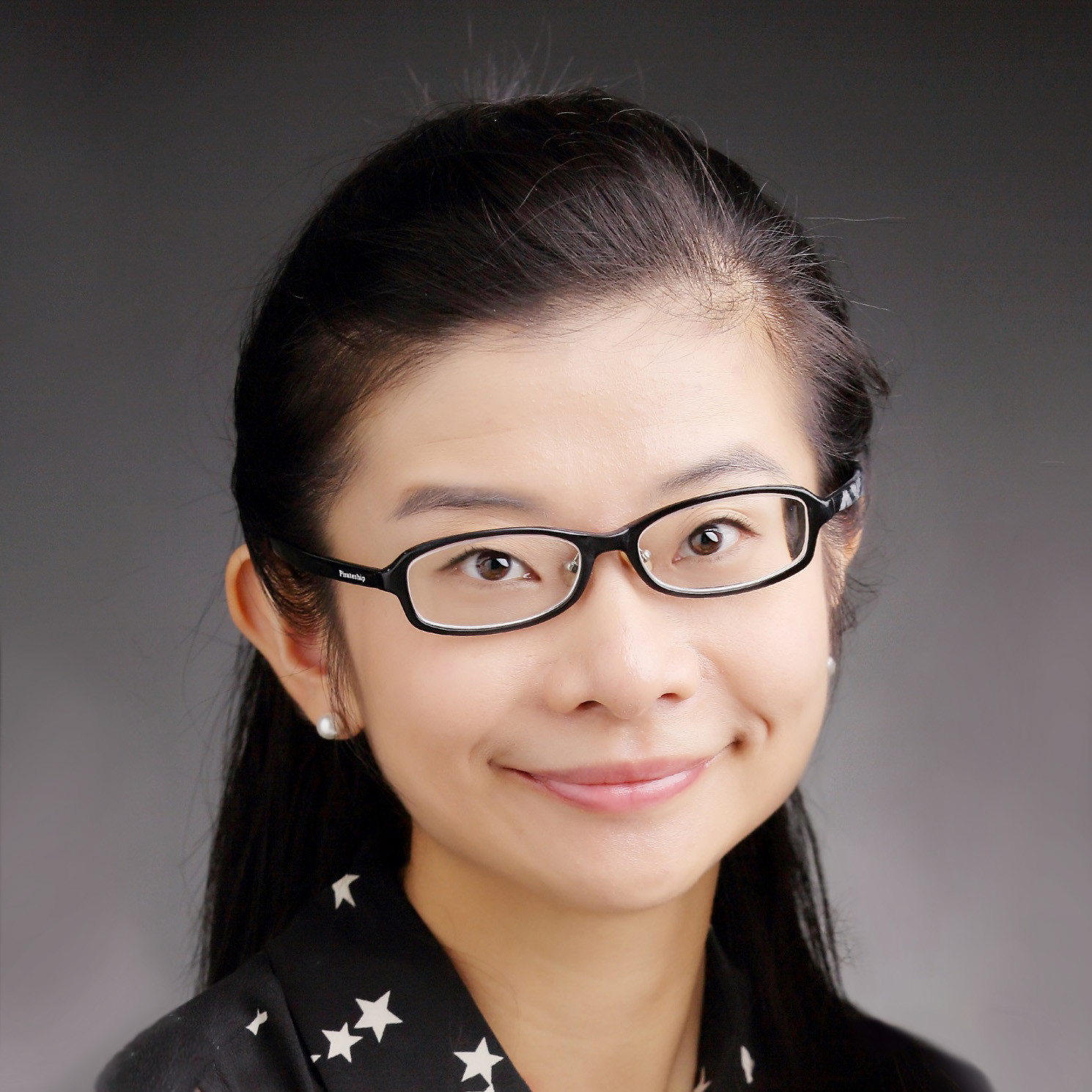}}]{Jiaying Liu (M’10-SM’17)} is currently an Associate Professor with the Wangxuan Institute of Computer Technology, Peking University. She received the Ph.D. degree (Hons.) in computer science from Peking University, Beijing China, 2010. She has authored over 100 technical articles in refereed journals and proceedings, and holds 42 granted patents. Her current research interests include multimedia signal processing, compression, and computer vision. Dr. Liu is a Senior Member of IEEE/CCF/CSIG. She was a Visiting Scholar with the University of Southern California, Los Angeles, from 2007 to 2008. She was a Visiting Researcher with the Microsoft Research Asia in 2015 supported by the Star Track Young Faculties Award. She has served as a member of Multimedia Systems \&amp; Applications Technical Committee (MSA-TC), Visual Signal Processing and Communications Technical Committee (VSPC) and Education and Outreach Technical Committee (EO-TC) in IEEE Circuits and Systems Society, a member of the Image, Video, and Multimedia (IVM) Technical Committee in APSIPA. She has served as the Associate Editor for IEEE Trans. on Image Processing, and Elsevier JVCI. She has also served as the Technical Program Chair of IEEE VCIP-2019/ACM ICMR-2021, the Publicity Chair of IEEE ICME-2020/ICIP-2019/VCIP-2018, and the Area Chair of ECCV-2020/ICCV-2019. She was the APSIPA Distinguished Lecturer (2016-2017).
\end{IEEEbiography}
\begin{IEEEbiography}[{\includegraphics[width=1in,clip,keepaspectratio]{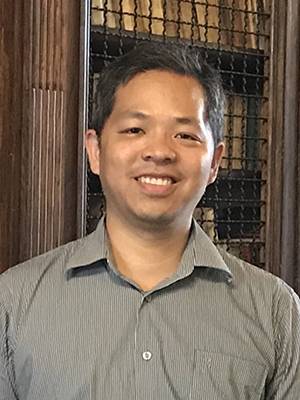}}]{Wen-Huang Cheng} is Professor with the Institute of Electronics, National Chiao Tung University (NCTU), Hsinchu, Taiwan. He is also Jointly Appointed Professor with the Artificial  Intelligence and Data Science Program, National Chung Hsing University (NCHU), Taichung, Taiwan. Before joining NCTU, he led the Multimedia Computing Research Group at the Research Center for Information Technology Innovation (CITI), Academia Sinica, Taipei, Taiwan, from 2010 to 2018. His current research interests include multimedia, artificial intelligence, computer vision, and machine learning. He has actively participated in international events and played important leading roles in prestigious journals and conferences and professional organizations, like Associate Editor for IEEE Transactions on Multimedia, General co-chair for IEEE ICME (2022) and ACM ICMR (2021), Chair-Elect for IEEE MSA technical committee, governing board member for IAPR. He has received numerous research and service awards, including the 2018 MSRA Collaborative Research Award, the 2017 Ta-Yu Wu Memorial Award from Taiwan’s Ministry of Science and Technology (the highest national research honor for young Taiwanese researchers under age 42), the 2017 Significant Research Achievements of Academia Sinica, the Top 10\% Paper Award from the 2015 IEEE MMSP, and the K. T. Li Young Researcher Award from the ACM Taipei/Taiwan Chapter in 2014. He is IET Fellow and ACM Distinguished Member.
\end{IEEEbiography}






\end{document}